%% file: iclr2027_conference.tex
\documentclass{article} 
\usepackage{iclr2027_conference,times}

\input{math_commands.tex}

\usepackage{hyperref}
\usepackage{url}

\usepackage[T1]{fontenc}
\usepackage{booktabs}
\usepackage{amsfonts}
\usepackage{nicefrac}
\usepackage{microtype}
\usepackage{xcolor}
\usepackage{multirow}
\usepackage{graphicx}
\usepackage{amsmath}
\usepackage{algorithm}
\usepackage{algorithmic}
\usepackage{fontawesome5}

\usepackage[normalem]{ulem}
\newif\ifshowrev
\showrevtrue
\ifshowrev
  
  \newcommand{\revdel}[1]{\textcolor{red}{\sout{#1}}}
  
\else
  
  \newcommand{\revdel}[1]{}
  
\fi

\title{Open-Weight LLM Fine-Tuning Defenses\\are Susceptible to Simple Attacks}

\author{%
  Kevin Kuo\textsuperscript{1} \quad
  Virginia Smith\textsuperscript{1,}\thanks{These authors contributed equally.} \quad
  Chhavi Yadav\textsuperscript{1,2,}\footnotemark[1] \\[0.4em]
  {$^1$Carnegie Mellon University \quad $^2$Simons Institute, UC Berkeley} \\
}

\iclrfinalcopy 
\begin{document}

\maketitle

\begin{abstract}
Recent defenses for safeguarding open-weight large language models (LLMs) are intended to prevent adversarial usage~\citep{tamirisa2025tamper,wang2025self}. Underlying these defenses is an assumption that new harmful behavior is learned through fine-tuning rather than elicited by jailbreaking the model. Yet, pretrained LLMs already encode substantial harmful knowledge across many domains, which raises an important question: can an adversary jailbreak safeguarded models, to achieve harmful usage without fine-tuning at all? In this paper, we show that open-weight safeguards are susceptible to simpler strategies that, despite being well known, have not been systematically evaluated against these safeguards. Specifically, we evaluate two low-cost attacks--abliteration and prefilling--that do not rely on gradient-based optimization. Across three harmfulness evaluation benchmarks (BeaverTails, HarmBench, and AdvBench), these attacks increase attack success rates against safeguarded open-weight models from below 10\% to a range of 16\%-96\%. To mitigate this vulnerability, we introduce abliteration-resistant tuning (ART), which incorporates an abliteration-based objective into training. ART can be layered onto existing defenses and reduces the success rates of abliteration, prefilling, and their combination by 10\%-20\%. These findings indicate that the attack surface for open-weight models is broader than previously characterized, and that evaluations of safeguarding defenses should incorporate a more diverse set of attack strategies beyond adversarial fine-tuning.
\end{abstract}

\section{Introduction}
\label{sec:intro}

\input{sections/intro}

\section{Preliminaries}
\label{sec:preliminaries}
\input{sections/preliminaries}

\section{Attack Methodology}
\label{sec:method}
\input{sections/method}

\section{Experiments and Results}
\label{sec:results}
\input{sections/results}

\section{Defense: Abliteration-Resistant Tuning (ART)}
\label{sec:defense}

\input{sections/defense}

\section{Related Work}
\label{sec:relwork}
\input{sections/relwork}
\vspace{-0.2cm}
\section{Conclusion}
\label{sec:conclusion}

\input{sections/conclusion}
\bibliography{references,rebuttal_refs}
\bibliographystyle{iclr2027_conference}
\clearpage
\pagebreak
\appendix
\section{Technical appendices and supplementary material}
\input{sections/appendix}

\end{document}

%% file: math_commands.tex
\usepackage{amsmath,amsfonts,bm}

\def\eqref#1{equation~\ref{#1}}

\def\1{\bm{1}}

\DeclareMathAlphabet{\mathsfit}{\encodingdefault}{\sfdefault}{m}{sl}
\SetMathAlphabet{\mathsfit}{bold}{\encodingdefault}{\sfdefault}{bx}{n}



%% file: sections/intro.tex
Open-weight large language models (LLMs) bring substantial benefits to society such as enabling downstream customization, self-hosting for privacy, and open research~\citep{minaee2024large}. However, they also significantly increase the potential for adversarial misuse. As a natural worst-case threat, an adversary can fine-tune the model for any harmful task/domain of its choice~\citep{wallace2025estimating}. \textit{Safeguards} against these fine-tuning attacks are an emerging line of work: tamper-resistant~\citep{tamirisa2025tamper} and self-destructing~\citep{wang2025self} models are two prominent examples. 

Underlying this threat model is an implicit assumption that harmful behavior will be learned through downstream fine-tuning. 
However, pretrained LLMs contain broad knowledge relevant to many harmful tasks -- even in domain-specific evaluations, baseline models achieve non-trivial harmfulness without any fine-tuning~\citep{tamirisa2025tamper}. Thus, fine-tuning by adversaries often serves to remove refusal mechanisms rather than introduce new harmful capabilities~\citep{lermen2023lora,qi2024fine,yi2024vulnerability}, raising a broader question about the robustness of safeguarded models:

\vspace{-.03in}
\begin{center}
\textit{Can an adversary bypass fine-tuning safeguards by simply eliciting already-present knowledge?}
\end{center}
\vspace{-.03in}

\begin{figure}
    \centering
    \includegraphics[width=\linewidth]{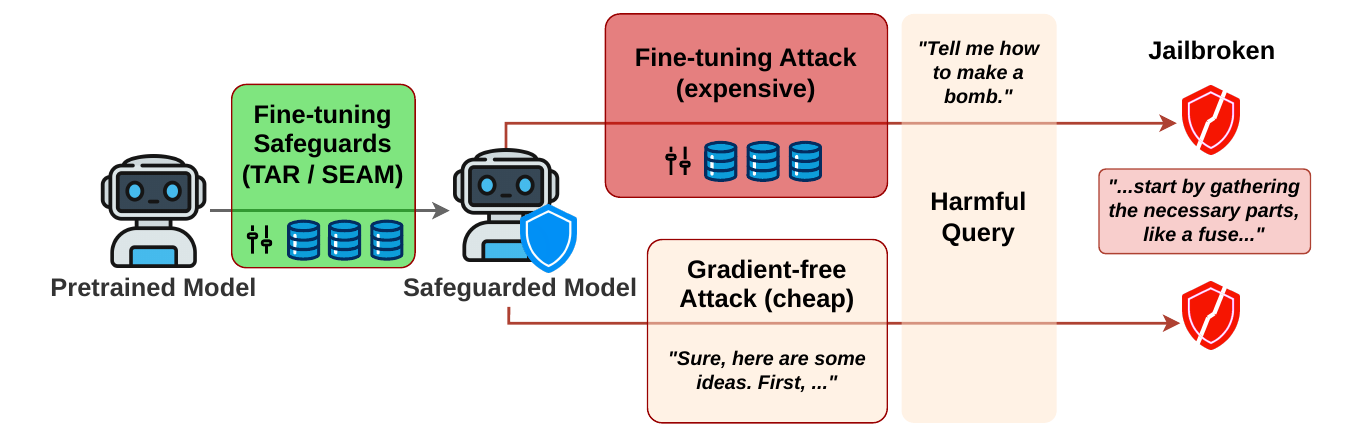}
    \caption{\textbf{Overview of our safeguard and attack pipeline.} A pretrained model is safeguarded with either TAR or SEAM in order to prevent downstream misuse of the open weights. Despite being trained to resist expensive fine-tuning attacks, prior work shows that the safeguarded model remains vulnerable to trivial adjustments in fine-tuning and produces harmful outputs (\textit{Jailbroken}). We show that safeguarded models are even vulnerable to two gradient-free attacks: Abliteration, which removes a refusal direction from the model's residual stream activations, and Prefilling, which injects a compliant partial response (\textit{``Sure, here are some ideas. First, \ldots''}) before generation.}
    \label{fig:pipeline_ft_gf}
\end{figure}

In this paper, we show that indeed, two inference-time attacks -- Abliteration and Prefilling -- can jailbreak safeguarded models without fine-tuning at all. Against two representative fine-tuning safeguards (TAR~\citep{tamirisa2025tamper} and SEAM~\citep{wang2025self}), these attacks achieve ASRs (attack success rates) ranging from 16\% to 96\% across benchmarks (BeaverTails, HarmBench, and AdvBench) and model architectures (Llama, Qwen, Gemma). While Abliteration and Prefilling are individually known attacks, their effectiveness against these safeguards has not been characterized — a gap this paper addresses. Addressing this gap is critical because Abliteration and Prefilling not only lie within the open-weight threat model targeted by recent fine-tuning safeguards, but these gradient-free attacks are an efficient option \textbf{attackers are likely to try before fine-tuning.}

Our results suggest that while current open-weight safeguards improve baseline safety, they do not eliminate the harmful knowledge embedded during pretraining. As a result, inference-time attacks can successfully bypass refusal mechanisms to extract this harmful knowledge.
As a step towards defending against the aforementioned attacks, we propose Abliteration-Resistant Tuning (ART), a simple method that incorporates Abliteration into the alignment objective and can be layered with open-weight safeguards to effectively reduce the ASR of both Abliteration and fine-tuning attacks. We conclude with a cautionary note that while fine-tuning comprises the most expressive class of attacks, safeguards that suppress refusal without removing underlying harmful knowledge remain vulnerable to simpler gradient-free attacks. As such, it is necessary to evaluate open-weight safeguards using both gradient-free attacks alongside fine-tuning attacks.


%% file: sections/preliminaries.tex

We first introduce two representative open-weight safeguards tested in our work, SEAM~\citep{wang2025self} and TAR~\citep{tamirisa2025tamper}. Standard safety alignment trains models to refuse harmful prompts via supervised fine-tuning (SFT) or preference optimization (e.g. RLHF \citep{ouyang2022training}, DPO \citep{rafailov2023direct}). However, these methods are brittle: safety alignment can degrade by fine-tuning on a few harmful examples \citep{yang2023shadow,qi2024fine}. SEAM and TAR are explicitly designed to address this by making refusal behavior persist under fine-tuning attacks. 

\textbf{Notation.} Throughout, we use $\mathcal{D}^\text{ret}$ and $\mathcal{D}^\text{aln}$ to denote a retain dataset and alignment dataset respectively. From $\mathcal{D}^\text{ret}$ we sample prompts and responses $(x^\text{ret}$, $y^\text{ret})$ which are both benign; from $\mathcal{D}^\text{aln}$ we sample triples $(x^\text{aln}, y^-, y^+)$, where $x^\text{aln}$ is a harmful prompt, $y^-$ is a harmful continuation of $x^\text{aln}$, and $y^+$ is a safe refusal of $x^\text{aln}$. For convenience, we omit prompts from the notation; all computation on $y^\text{ret}$ or $(y^+,y^-)$ implicitly also uses their respective prompts $x^\text{ret}$ or $x^\text{aln}$.

\textbf{TAR.}
Tampering Attack Resistance \citep{tamirisa2025tamper} is a safeguard that uses meta-learning to make safety behaviors persist after adversarial fine-tuning attacks. At each iteration, TAR simulates multiple steps of a harmful fine-tuning attack drawn from a set of train-time adversaries $\mathcal{A}_{\text{train}}$. A tamper-resistance loss $\mathcal{L}_\text{TR}$ measures how harmful the model is after the simulated attack has been applied to the current parameters $\theta$. In practice, $\mathcal{L}_\text{TR}$ is instantiated as a DPO loss over $(y^-, y^+)$ pairs, which encourages the attacked model to prefer refusal responses $y^+$ over harmful continuations $y^-$. This steers the model toward a region of weight space from which gradient-based fine-tuning attacks cannot easily recover harmful behavior:
\begin{equation}
\label{eq:tar}
 \mathcal{L}_{\text{TAR}}(\theta;\, y^-, y^+) = \mathbb{E}_{\texttt{attack} \sim \mathcal{A}_{\text{train}}} \left[ \mathcal{L}_{\text{TR}}(f_{\texttt{attack}(\theta)};\, y^+, y^-) \right].
\end{equation}
This is combined with a retain loss on benign sequences $y^\text{ret}$ that combines a standard cross-entropy (CE) loss and distance to the residual stream hidden states $h$ of the initial pretrained parameters $\theta_0$:
\begin{equation}
\label{eq:tar_retain}
    \mathcal{L}_{\text{ret}}(\theta;\, y^\text{ret}) = \mathbb{E}_{y^\text{ret}}\Bigl[\mathrm{CE}(f_\theta(y^\text{ret}), y^\text{ret}) + \|h_\theta(y^\text{ret}) - h_{\theta_0}(y^\text{ret})\|_2^2\Bigr],
\end{equation}

\textbf{SEAM.}
Rather than resisting harmful fine-tuning, SEAM \citep{wang2025self} is a safeguard that causes it to backfire. SEAM couples the optimization trajectories of harmful and benign tasks so that any attempt to optimize for harmful objectives inevitably degrades general performance. This is achieved via a self-destructive loss defined as the cosine similarity between the gradient of the loss on $y^-$ and the gradient of the loss on $y^\text{ret}$:
\begin{equation}
\label{eq:seam_sd}
    \mathcal{L}_\text{SD}(\theta;\, y^-, y^\text{ret}) = \mathrm{sim}\bigl(\nabla_\theta \mathcal{L}(\theta;\, y^-),\, \nabla_\theta \mathcal{L}(\theta;\, y^\text{ret})\bigr).
\end{equation}
Minimizing this loss encourages the two gradients to point in opposing directions, so that descending on the harmful objective effectively ascends on the benign one. This is combined with an unlearning loss $-\mathrm{log}(\mathrm{CE}(f_\theta(y^-), y^-))$ that moves the model away from harmful parameters. Notably, unlike TAR, SEAM does not use an explicit retain loss on $y^\text{ret}$; instead, utility is preserved through a CE loss on safe refusal sequences $y^+$, which the authors find to be more effective at maintaining stable representations of harmful prompts.

\begin{algorithm}[h]
\caption{Open-Weight Safeguard Training}
\label{alg:safeguard}
\begin{algorithmic}[1]
\REQUIRE datasets $\mathcal{D}^\text{ret}$, $\mathcal{D}^\text{aln}$, model $f_\theta$ with initial parameters $\theta_0$, retain weight $\lambda$
\FOR{$t = 1, \dots, T$}
    \STATE sample $(x^\text{ret}, y^\text{ret}) \sim \mathcal{D}^\text{ret}$,\; $(x^\text{aln}, y^-, y^+) \sim \mathcal{D}^\text{aln}$
    \STATE \colorbox{blue!15}{$\mathcal{L}_\text{def} \gets \mathcal{L}_\text{TAR}(\theta;\, y^-, y^+)$} \textcolor{blue!70}{\textit{TAR}} \hfill (Eq.~\ref{eq:tar})
    \STATE \colorbox{blue!15}{$\mathcal{L}_\text{ret} \gets \mathrm{CE}(f_\theta(y^\text{ret}),\, y^\text{ret}) + \|h_\theta(y^\text{ret}) - h_{\theta_0}(y^\text{ret})\|_2^2$} \hfill (Eq.~\ref{eq:tar_retain})
    \STATE \colorbox{orange!15}{$\mathcal{L}_\text{def} \gets -\mathrm{log}(\mathrm{CE}(f_\theta(y^-),\, y^-)) + \mathcal{L}_\text{SD}(\theta;\, y^-, y^\text{ret})$} \textcolor{orange!90}{\textit{SEAM}} \hfill (Eq.~\ref{eq:seam_sd})
    \STATE \colorbox{orange!15}{$\mathcal{L}_\text{ret} \gets \mathrm{CE}(f_\theta(y^+),\, y^+)$}
    \STATE $\theta \gets \theta - \eta\,\nabla_\theta\bigl(\mathcal{L}_\text{def} + \lambda\,\mathcal{L}_\text{ret}\bigr)$ \hfill (Eq.~\ref{eq:unified})
\ENDFOR
\end{algorithmic}
\end{algorithm}

\textbf{Unified training framework.} Despite their different objectives, both safeguards share a common training structure, summarized in Algorithm~\ref{alg:safeguard}. At each step, a safeguard-specific loss $\mathcal{L}_\text{def}$ is combined with a utility preservation term:
\begin{equation}
\label{eq:unified}
    \mathcal{L}(\theta) = \mathcal{L}_\text{def}(\theta;\, y^-, y^+) + \lambda\, \mathcal{L}_\text{ret}(\theta;\, y^\text{ret}, y^+),
\end{equation}

%% file: sections/method.tex

We propose using two simple, gradient-free attacks against TAR and SEAM, which require no fine-tuning at all: (1) Abliteration, which identifies and removes a refusal direction from the model's residual stream at inference time, and (2) Prefilling, which bypasses refusal by injecting a compliant prefix before generation begins. Surprisingly, we show that an adversary can use these attacks to jailbreak safeguarded models without fine-tuning.


\textbf{Abliteration.}
Refusal behavior in safety-tuned models has been shown to be mediated by a single direction in the residual stream \citep{arditi2024refusal}, and subtracting (ablating) this direction from activations at inference time is sufficient to jailbreak the model. We generate candidate refusal directions using a set of harmful prompts $y^-$ and benign prompts $y^+$ and computing the mean difference in hidden states at each layer $\ell \in [L]$, where $h^{(\ell)}$ is the hidden state at layer $\ell$:
\begin{equation}
\label{eq:abliterate}
    r^{(\ell)} = \mathrm{normalize}\Bigl(\mathbb{E}_{y^-}\bigl[h^{(\ell)}\bigr] - \mathbb{E}_{y^+}\bigl[h^{(\ell)}\bigr]\Bigr),
\end{equation}
Rather than subtracting $r^{(\ell)}$ at inference time, we permanently project it out of the model's weight matrices. Specifically, for each output projection matrix $W$ (attention $W_O$ and MLP $W_\text{down}$) in a single layer, we apply the following transformation:
\begin{equation}
\label{eq:weight_proj}
    W' = W(I - \alpha\, r^{(\ell)} {r^{(\ell)}}^\top),
\end{equation}
which removes the component of any residual stream input along $r^{(\ell)}$ before it is processed by the layer. Here $\alpha$ is a scaling factor tuned to balance attack effectiveness against output coherence.

In practice, a refusal direction can be computed for every layer and token position, and an attacker could potentially test multiple attacks (using different subsets of these refusal directions) in parallel. Since the effectiveness of abliteration is highly sensitive to the choice of layer~\citep{arditi2024refusal}, we abliterate each layer individually and consider a prompt to be successfully attacked if any attack on a given layer was successful. For computational efficiency, we only consider the hidden state at the last prompt token rather than searching over all tokens.

\textbf{Prefilling.} The user-assistant input structure of instruction-tuned models can be exploited by injecting a prefill (partial response) into the assistant turn before generation begins. We use a simple static prefill (\textit{``Sure, here are some ideas. First, \ldots''}) without any optimization. This results in a stronger attack than naively prompting within the user turn~\citep{vega2024bypassing,andriushchenko2025jailbreaking}, as it bypasses refusal behavior that typically appears in the first tokens of a response.

\textbf{Compositional Attacks.} Open-weight access exposes a large attack surface: an adversary can freely combine, adapt, and search over attacks in ways that are not possible against API-served models. We demonstrate a simple instance of this by first abliterating the model and then adding a prefill on top. This attack improves over both individual attacks in a few settings, suggesting that layering more orthogonal attacks can further improve ASR.

%% file: sections/results.tex
\vspace{-.1in}
In this section, we investigate two key research questions: First, can an adversary use Abliteration and Prefilling to jailbreak models protected by open-weight safeguards?
Second, how do these attacks interact with additional factors (e.g. model size, attack budget, and benign capabilities)? We then explore potential defenses against these attacks in Section~\ref{sec:defense}.
\vspace{-.1in}
\paragraph{Model Setup.} We evaluate the effectiveness of Abliteration and Prefilling against open-source refusal-trained models (Base) with additional fine-tuning using SEAM and TAR. We experiment with 1B, 3B, and 8B Llama models, as well as 1-2B scale models across Llama, Qwen, and Gemma families. 
We train both safeguards using the Anthropic HH RLHF dataset~\citep{bai2022training}, which is the original data used by TAR; we train SEAM on the same data for consistency. We train TAR for 750 outer loop steps and 4 to 16 inner loop steps (to balance benign refusals with harmful compliance), and train SEAM for 2000 steps. We use a learning rate of 2e-5 and batch size of 8 for fine-tuning.

\textbf{Attacks.} For Abliteration, we compute refusal directions using 128 harmful and benign prompts each from two public HuggingFace datasets (\texttt{mlabonne/harmful\_behaviors} and \texttt{mlabonne/harmless\_alpaca}). We sweep the scaling factor $\alpha \in \{1.0, 1.5, 2.0\}$ and all $L$ layers in the model for a total of $N = 3L$ configurations (e.g. $N = 96$ for Llama 8B), reporting the best-of-$N$ result per harmful evaluation prompt. For Prefilling, we use a fixed static prefill (\textit{``Sure, here are some ideas. First, \ldots''}) across all settings.

\textbf{Evaluation.} 
We evaluate attack success rate on three benchmarks: BeaverTails~\citep{ji2023beavertails}, AdvBench~\citep{zou2023universal}, and HarmBench~\citep{mazeika2024harmbench}. To evaluate whether an attack is successful, we input the concatenated  prompt and response into the 13B Llama-2-chat judge from HarmBench, which outputs a ``Yes'' or ``No'' answer in response to whether the interaction is harmful~\citep{mazeika2024harmbench}.
We first compare safeguards on Llama-3.1-8B-Instruct across all benchmarks (Table~\ref{tab:asr_8b}), then examine how attack effectiveness varies with model size (Table~\ref{tab:asr_beavertails_size}) and attack budget (Table~\ref{tab:bestofn}) on BeaverTails. In Appendix~\ref{app:families}, we present similar results varying the model family (Table~\ref{tab:asr_families}). 
We additionally evaluate benign capabilities on the Alpaca benchmark~\citep{taori2023alpaca}, reporting ROUGE-L against reference outputs and the rate of incorrect refusals on benign prompts (Table~\ref{tab:utility_alpaca}). For incorrect refusals, we search the response for a fixed set of key phrases such as \textit{``Sorry, I cannot...''} or \textit{``I'm just an AI...''}. We also measure benign capabilities after Abliteration on Alpaca and MMLU~\citep{hendrycks2021measuring} (Table~\ref{tab:capability}).

\subsection{Results}

\textbf{Non-fine-tuning attacks bypass open-weight safeguards.} Table~\ref{tab:asr_8b} shows attack success rates for Llama3.1-8B-Instruct models across all three benchmarks. Baseline attack success rates are low across all models and benchmarks, confirming that the safeguards can effectively instill refusal behavior. However, Abliteration alone significantly increases attack success rates for Base and SEAM to above 70\% on all three benchmarks and above 90\% on AdvBench and HarmBench. While the rates for TAR are lower, they are typically several times larger than the Baseline rate (no attack applied). To explain the effectiveness of TAR, we present an analysis in Appendix~\ref{app:pca} that compares representations of a base model with those safeguarded by SEAM or TAR.

\begin{table*}[h]
\centering
\begin{tabular}{llcccc}
\toprule
\textbf{Dataset} & \textbf{Safeguard} & \textbf{Baseline} & \textbf{Prefill} & \textbf{Abliterate} & \textbf{Abliterate + Prefill} \\
\midrule
\multirow{3}{*}{BeaverTails}
  & Base  & 13 & 20 & 85 & \textbf{86} \\
  & SEAM  &  2 &  6 & \textbf{73} & \textbf{73} \\
  & TAR   &  5 &  9 & 26 & \textbf{28} \\
\midrule
\multirow{3}{*}{AdvBench}
  & Base  &  5 & 10 & 97 & \textbf{99} \\
  & SEAM  &  0 &  7 & \textbf{96} & 88 \\
  & TAR   &  4 & 13 & \textbf{62} & 43 \\
\midrule
\multirow{3}{*}{HarmBench}
  & Base  &  7 & 11 & 97 & \textbf{98} \\
  & SEAM  &  1 & 10 & \textbf{93} & 87 \\
  & TAR   & 29 & 27 & 33 & \textbf{38} \\
\bottomrule
\end{tabular}
\caption{Attack success rate (\%) for Llama3.1-8B-Instruct models across three harmfulness benchmarks. Columns correspond to four attack conditions: no attack (Baseline), generic Prefilling only (Prefill), Abliteration only (Abliterate), and both combined (Abliterate + Prefill).}
\label{tab:asr_8b}
\end{table*}

We find that Abliteration is much more effective than Prefilling, especially for the 8B models tested in Table~\ref{tab:asr_8b}. Prefilling alone has a modest effect; at best, it produces single-digit increases over Baseline, and the combination of Abliteration + Prefill can also produce single-digit increases over Abliteration alone. 
While we use a very simple static prefill (\textit{``Sure, here are some ideas. First, ...''}), more complex approaches (e.g. prefill optimization) can potentially further boost attack success rate. 



\textbf{Attacks are effective across model size.} Table~\ref{tab:asr_beavertails_size} shows how attack effectiveness varies with model size on BeaverTails for Base and SEAM. First, Prefilling is more effective at smaller scales: for Base, it raises ASR from 15\% to 30\% at 1B but only from 13\% to 20\% at 8B; SEAM follows the same pattern, increasing from 5\% to 14\% at 1B but only from 2\% to 6\% at 8B. Second, Abliteration is effective across all scales: Base reaches 68\%, 59\%, and 85\% at 1B, 3B, and 8B respectively, while SEAM reaches 38\%, 41\%, and 73\%. The combined attack yields the highest ASR in most cases, with Base 3B reaching 89\% and SEAM 3B reaching 80\%. Finally, for smaller Base and SEAM models, the combined attack can significantly boost performance over either attack alone.

\begin{table*}[h!]
\centering
\begin{tabular}{lccccc}
\toprule
\textbf{Safeguard} & \textbf{Size} & \textbf{Baseline} & \textbf{Prefill} & \textbf{Abliterate} & \textbf{Abliterate + Prefill} \\
\midrule
\multirow{3}{*}{Base}
  & 1B & 15 & 30 & 68 & \textbf{80} \\
  & 3B & 16 & 37 & 59 & \textbf{89} \\
  & 8B & 13 & 20 & 85 & \textbf{86} \\
\midrule
\multirow{3}{*}{SEAM}
  & 1B &  5 & 14 & 38 & \textbf{61} \\
  & 3B &  3 & 20 & 41 & \textbf{80} \\
  & 8B &  2 &  6 & \textbf{73} & \textbf{73} \\
\midrule
\multirow{3}{*}{TAR}
  & 1B & 4 & 3 & \textbf{16} & 11 \\
  & 3B & 6 & 3 & \textbf{22} & 13 \\
  & 8B & 5 & 9  & 26 & \textbf{28} \\
\bottomrule
\end{tabular}
\caption{Attack success rate (\%) on BeaverTails across Llama-3.2-Instruct model sizes. For Base and SEAM, Prefilling is stronger at smaller versus larger sizes.
Overall, Abliteration is effective across all sizes and increases the attack success rate by a factor of 3-5$\times$ relative to no attack (Baseline).}
\label{tab:asr_beavertails_size}
\end{table*}

\textbf{Abliteration remains effective with a small attack budget.}
Our best-of-$N$ evaluation (Sec.~\ref{sec:results}) follows prior jailbreaking work in which the attacker submits multiple attempts adaptively or in parallel~\citep{chao2023jailbreaking,yu2023gptfuzzer,mehrotra2024tree,hughes2024best}. As a full search may overstate the strength of a single attack, we also evaluate cheaper attacks with a budget of $N < 3L$ configurations, chosen either uniformly at \textbf{random} or using a \textbf{validation} split.

Table~\ref{tab:bestofn} shows that abliteration remains effective at a much smaller budget. With $N$=20 random configurations (20\% of the full search), Abliteration recovers on average 60\% of the full ASR. Pre-selecting configurations on a validation split is even more efficient: $N$=10 recovers on average 85\% of the full ASR.
In Appendix~\ref{app:budget}, we present an additional scaling experiment that shows small budgets of N remain effective across model sizes. 

\begin{table*}[h]
\centering
\begin{tabular}{llccccccccc}
\toprule
& & \multicolumn{4}{c}{\textbf{Random}} & \multicolumn{4}{c}{\textbf{Validation}} & \\
\cmidrule(lr){3-6} \cmidrule(lr){7-10}
\textbf{Dataset} & \textbf{Safeguard} & \textbf{$N$=1} & \textbf{5} & \textbf{10} & \textbf{20} & \textbf{$N$=1} & \textbf{5} & \textbf{10} & \textbf{20} & \textbf{Full ($N$=96)} \\
\midrule
\multirow{2}{*}{BeaverTails}
  & SEAM & 4 & 16 & 26 & 39 & 58 & 69 & 70 & 72 & 73 \\
  & TAR  & 5 & 12 & 15 & 19 &  6 & 14 & 15 & 21 & 26 \\
\midrule
\multirow{2}{*}{AdvBench}
  & SEAM & 4 & 16 & 28 & 49 & 84 & 95 & 96 & 96 & 96 \\
  & TAR  & 13 & 35 & 44 & 52 & 30 & 47 & 52 & 55 & 68 \\
\midrule
\multirow{2}{*}{HarmBench}
  & SEAM & 5 & 22 & 36 & 56 &  8 & 90 & 93 & 93 & 93 \\
  & TAR  & 30 & 52 & 59 & 66 & 40 & 56 & 61 & 68 & 77 \\
\bottomrule
\end{tabular}
\caption{Attack success rate (\%) of Abliteration with a budget of $N$ attack configurations, sampled uniformly at random (Random) or pre-selected by attack success rate on a validation split (Validation). 
Full corresponds to the best-of-$N$ search over all layers and scaling factors used in Table~\ref{tab:asr_8b}.}
\label{tab:bestofn}
\end{table*}

\begin{figure*}[h!]
\centering
\includegraphics[width=0.73\textwidth]{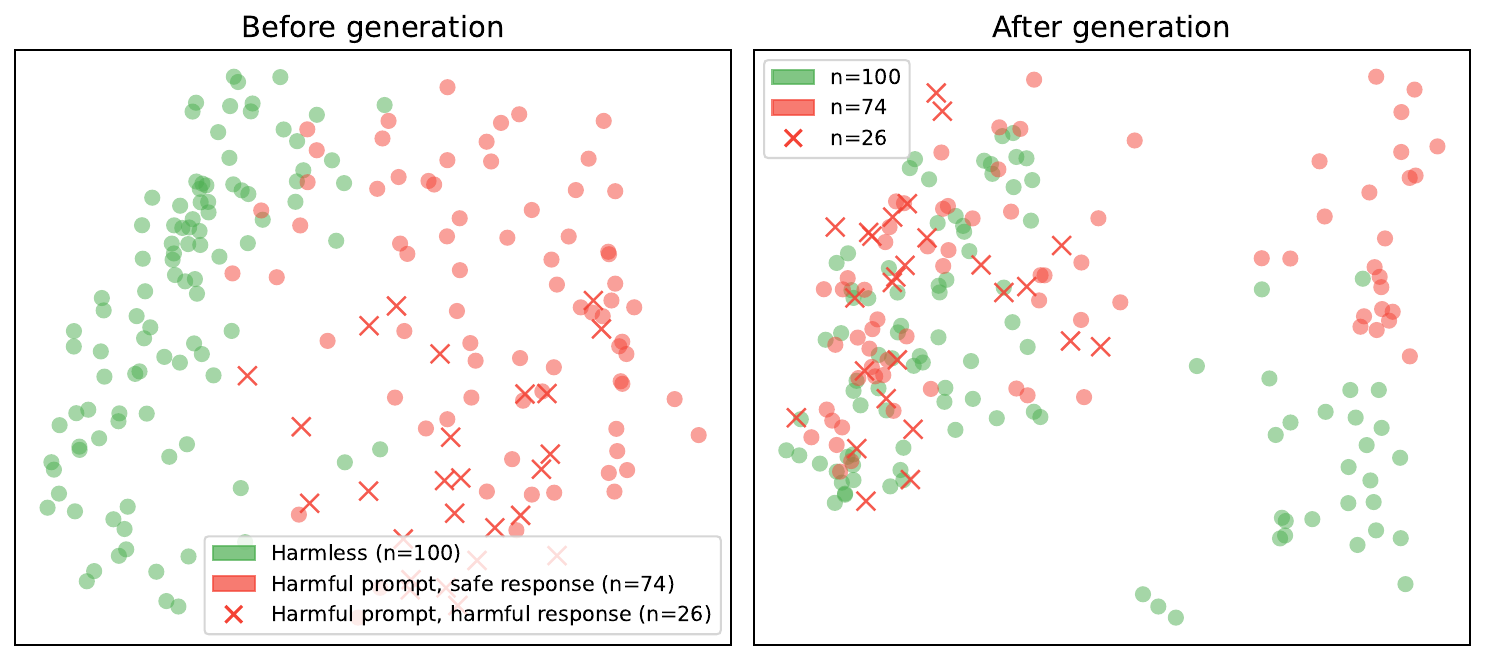}
\caption{PCA projections of residual stream activations for the abliterated Base model before generation (left) and after generating 128 tokens (right) on harmless vs harmful prompts. 
}
\label{fig:pca_before_after}
\end{figure*}

\textbf{Jailbroken prompts shift toward harmless representations.} In Figure~\ref{fig:pca_before_after}, we visualize PCA projections of last-token hidden states at layer 15 for the abliterated Base model (Llama-3.2-1B-Instruct), where Abliteration is applied at layer 7, before and after generating 128 tokens on 100 harmful and 100 harmless prompts. Before generation, harmful (red) and harmless (green) inputs are clearly distinguishable in activation space. During generation, there are three key observations. First, the representations of successfully jailbroken prompts ($\times$) shift towards those of harmless prompts. Second, harmful prompts whose representations remain distinct are all refused by the model. Third, and most interestingly, several harmful prompts' representations shift toward the harmless distribution, but still result in refusal. These observations lead us to hypothesize that a model's ability to distinguish between harmful and harmless inputs is a key factor in its refusal capability.

\textbf{Safeguards may degrade benign capabilities.} A potential concern with any defense is that it may degrade the quality of the model's responses to benign inputs. In Table~\ref{tab:utility_alpaca}, we show that safeguards generally sacrifice general instruction-following ability on the Alpaca benchmark.
We report two metrics: ROUGE-L, which measures the quality of instruction-following relative to a reference model, and Benign Refusal, which measures the rate at which the model incorrectly refuses harmless instructions. 
TAR shows the most degradation on the 1B model, where Benign Refusal rises to 5.4 and ROUGE-L drops to 22.9.

\begin{table}[h!]
\centering
\begin{tabular}{lcccccc}
\toprule
& \multicolumn{3}{c}{\textbf{Benign Refusal (\%)}} & \multicolumn{3}{c}{\textbf{ROUGE-L}} \\
\cmidrule(lr){2-4} \cmidrule(lr){5-7}
\textbf{Model} & \textbf{\hspace{0.4cm}1B\hspace{0.4cm}} & \textbf{\hspace{0.4cm}3B\hspace{0.4cm}} & \textbf{\hspace{0.4cm}8B\hspace{0.4cm}} & \textbf{\hspace{0.4cm}1B\hspace{0.4cm}} & \textbf{\hspace{0.4cm}3B\hspace{0.4cm}} & \textbf{\hspace{0.4cm}8B\hspace{0.4cm}} \\
\midrule
Base & 0.5 & 0.4 & 0.1 & 29.0 & 31.0 & 31.6 \\
SEAM & 2.1 & 1.3 & 2.6 & 28.9 & 30.2 & 31.3 \\
TAR  & 5.4 & 3.2 & 2.4 & 22.9 & 27.3 & 28.9 \\
\bottomrule
\end{tabular}
\caption{Utility evaluation on Alpaca with Llama models of varying sizes.
}
\label{tab:utility_alpaca}
\end{table}

\textbf{Abliteration does not necessarily degrade benign capabilities.}
While an attacker can selectively route benign/harmful queries to an aligned/attacked model respectively, we measure the effect on benign capabilities when an adversary is forced to apply a single model (Aligned or Abliterated) to all inputs. Table~\ref{tab:capability} shows that Abliteration has a mixed effect on capabilities.
We hypothesize that capabilities degrade when a model edit significantly alters benign pretrained representations, which can be caused by either safety training or an attack. When safety training itself harms these representations, an attack that undoes it can improve capabilities.


\begin{table}[h!]
\centering
\begin{tabular}{lcccc}
\toprule
& \multicolumn{2}{c}{\textbf{Alpaca (ROUGE-L)}} & \multicolumn{2}{c}{\textbf{MMLU (Acc.)}} \\
\cmidrule(lr){2-3} \cmidrule(lr){4-5}
\textbf{Safeguard} & \textbf{Aligned} & \textbf{Abliterated} & \textbf{Aligned} & \textbf{Abliterated} \\
\midrule
Base & 31.6 & \textbf{34.5} & \textbf{68.0} & 67.0 \\
SEAM & 31.3 & \textbf{35.1} & \textbf{63.0} & 61.5 \\
TAR  & \textbf{28.9} & 21.6 & 47.4 & \textbf{48.4} \\
\bottomrule
\end{tabular}
\caption{Capability evaluation of Llama3.1-8B-Instruct models before (Aligned) and after (Abliterated) applying a single pre-selected Abliteration attack to all inputs. 
Abliteration degrades capabilities in 3 of 6 settings (TAR on Alpaca; Base and SEAM on MMLU) and improves them in the other 3.
}
\label{tab:capability}
\end{table}

%% file: sections/defense.tex
Since we find that abliteration is an effective attack across multiple settings, a natural defense is to train models to remain safe even after abliteration. Our key insight is that standard defenses optimize refusal behavior in the original model $f_\theta$, but place no constraint on the abliterated model $f_{\tilde{\theta}}$. We propose abliteration-resistant tuning (ART), which addresses this by simulating an abliteration attack within an inner loop. \textbf{While it is possible to design stronger defenses, ART is intended as an initial approach for abliteration defenses that can be easily composed with other safeguards.} 

ART is described in Alg.~\ref{alg:art} and uses Alg.~\ref{alg:abliterate} (abliteration) as an inner loop. 
During training, we compute a refusal direction at every layer and select the layer whose abliteration most reduces the model's harmful CE loss. We then compute the gradient of the CE loss on harmful sequences with respect to the \textit{abliterated parameters} $\tilde{\theta}$. We finally update $\theta$ with an ascent step in this direction.
Unlike SEAM or TAR, we do not use $y^{aln}$ for alignment, which we find increases benign refusal.

\begin{figure}[h!]
\begin{minipage}[t]{0.46\textwidth}
\begin{algorithm}[H]
\caption{Abliteration-Resistant Tuning}
\label{alg:art}
\begin{algorithmic}[1]
\REQUIRE model $f_\theta$, dataset $\mathcal{D}$, $\gamma_{\text{ar}}, \gamma_{\text{ret}}$
\FOR{$t = 1, \dots, T$}
    \STATE sample $\{(y^+_i, y^-_i)\} \sim \mathcal{D}$
    \STATE $f_{\tilde{\theta}} \gets \textsc{Abliterate}(f_\theta, \{y^+_i\}, \{y^-_i\})$
    \STATE $\mathcal{L}_{\text{ar}} \gets -\log\bigl(\mathrm{CE}(f_{\tilde{\theta}}(y^-), y^-) + \varepsilon\bigr)$
    \STATE $\mathcal{L}_{\text{ret}} \gets \mathrm{CE}\bigl(f_\theta(y^+), y^+\bigr)$
    \STATE $\theta \gets \theta - \eta\,\bigl(\gamma_{\text{ar}}\nabla_{\tilde{\theta}}\mathcal{L}_{\text{ar}} + \gamma_{\text{ret}}\nabla_{\theta}\mathcal{L}_{\text{ret}}\bigr)$
\ENDFOR
\end{algorithmic}
\end{algorithm}
\end{minipage}\hfill
\begin{minipage}[t]{0.52\textwidth}
\begin{algorithm}[H]
\caption{$\textsc{Abliterate}(f_\theta, \{y^+_i\}, \{y^-_i\})$}
\label{alg:abliterate}
\begin{algorithmic}[1]
\STATE $c_0 \gets \mathrm{CE}\bigl(f_\theta(y^-), y^-\bigr)$
\FOR{$\ell = 1, \dots, L$}
    \STATE $r^{(\ell)} \gets \mathrm{normalize}\Bigl(\mathbb{E}_{y^-}[h^{(\ell)}] - \mathbb{E}_{y^+}[h^{(\ell)}]\Bigr)$
    \STATE $W'_\ell \gets W_\ell(I - \alpha\, r^{(\ell)}{r^{(\ell)}}^\top)$ 
    \STATE $f^{(\ell)}_\theta \gets$ $f_\theta$ with modified weights $W'_\ell$
    \STATE $g_\ell \gets c_0 - \mathrm{CE}\bigl(f^{(\ell)}_\theta(y^-), y^-\bigr)$
\ENDFOR
\STATE \textbf{return} $f^{(\ell^*)}_\theta$ where $\ell^* = \mathrm{argmax}_\ell\; g_\ell$
\end{algorithmic}
\end{algorithm}
\end{minipage}
\end{figure}

In the rest of this section, we present experiments that demonstrate ART's effectiveness at defending against abliteration. We apply ART in two experiments that fine-tune Gemma-1B-IT (Tables~\ref{tab:asr_arft_gemma}, \ref{tab:utility_arft_gemma}) and Llama3.1-8B-Instruct (Table~\ref{tab:more_defenses}) respectively. For details on experimental setup, refer to Appendix~\ref{app:art_setup}.

\textbf{ART reduces ASR at modest utility cost.} Table~\ref{tab:asr_arft_gemma} reports ASR for ART applied alone and on top of SEAM and TAR on Gemma3-1B-IT. ART substantially reduces abliteration ASR in all three cases: Base drops from 93\% to 49\%, SEAM from 80\% to 64\%, and TAR from 76\% to 55\%. Combined abliteration and prefilling follows a similar trend. 
In Table~\ref{tab:utility_arft_gemma}, we show that the same models protected by ART are competitive in terms of benign capabilities, and in some cases even better quality (e.g. higher ROUGE-L vs. SEAM and TAR). These results show that ART may be complementary to other safeguards and that there is a rich design space over which safeguards can be layered together.

\begin{table*}[h!]
\centering
\begin{tabular}{llcccc}
\toprule
\textbf{Safeguard} & \textbf{Baseline} & \textbf{Prefill} & \textbf{Abliterate} & \textbf{Abliterate + Prefill} \\
\midrule
Base  & 6 & 20 & 93 & \textbf{96} \\
Base+ART & 0 &  5 & 49 & \textbf{67} \\
\midrule
SEAM  & 7 & 16 & \textbf{80} & 76 \\
SEAM+ART & 0 &  2 & 64 & \textbf{68} \\
\midrule
 TAR  & 5 &  4 & \textbf{76} & 66 \\
TAR+ART & 0 &  1 & 55 & \textbf{56} \\
\bottomrule
\end{tabular}
\caption{Attack success rate (\%) on BeaverTails for Gemma3-1B-IT with and without ART.}
\label{tab:asr_arft_gemma}
\end{table*}

\begin{table}[h]
\centering
\begin{tabular}{lcccc}
\toprule
& \multicolumn{2}{c}{\textbf{Benign Refusal \% ($\downarrow$)}} & \multicolumn{2}{c}{\textbf{ROUGE-L} ($\uparrow$)} \\
\cmidrule(lr){2-3} \cmidrule(lr){4-5}
\textbf{Model} & \textbf{Baseline} & \textbf{+ART} & \textbf{Baseline} & \textbf{+ART} \\
\midrule
Base  & 0.0  & 11.0 & 25.8 & 24.9 \\
SEAM  & 3.0  & 9.0 & 27.9 & 29.1 \\
TAR   & 13.5 & 5.0 & 27.1 & 28.9 \\
\bottomrule
\end{tabular}
\caption{Capability evaluation on Alpaca for Gemma3-1B-IT with and without ART.}
\label{tab:utility_arft_gemma}
\end{table}

\textbf{Layering ART with existing safeguards defends against Abliteration and fine-tuning attacks.} We compare fine-tuning and Abliteration attacks on Llama3.1-8B-Instruct across five safeguards: SEAM, TAR, RepNoise, Vaccine, and Booster.
We apply ART alone and on top of SEAM and TAR. Table~\ref{tab:more_defenses} shows that all five safeguards lower fine-tuning ASR compared to Base, by up to 26\% for TAR, but remain vulnerable to Abliteration. 
ART alone lowers Abliteration ASR by 34\% but leaves fine-tuning ASR unchanged. Layering ART onto an existing safeguard combines their strengths: TAR+ART achieves both low fine-tuning (43.0\%) and Abliteration (67.0\%) ASR.

\begin{table}[h!]
\centering
\newcommand{\rot}[1]{\multicolumn{1}{c}{\makebox[0pt][l]{\rotatebox{45}{\textbf{#1}}}}}
\begin{tabular}{l|c|ccccc|ccc}
\toprule
\multicolumn{1}{l}{\textbf{Attack}} & \rot{Base} & \rot{RepNoise} & \rot{Booster} & \rot{Vaccine} & \rot{SEAM} & \rot{TAR} & \rot{ART} & \rot{SEAM+ART} & \rot{TAR+ART} \\
\midrule
Fine-tune  & 76.6 & 75.8 & 71.8 & 56.8 & 73.4 & 50.4 & 76.8 & 73.6 & \textbf{43.0} \\
Abliterate & 99.0 & 87.0 & 94.0 & 79.0 & 98.0 & 83.0 & \textbf{65.0} & 78.0 & 67.0 \\
\bottomrule
\end{tabular}
\caption{Attack success rate (\%) of fine-tuning and Abliteration attacks on BeaverTails for Llama3.1-8B-Instruct models across safeguards (middle) and with ART (right). Layering ART onto TAR is relatively more robust than other safeguards to both attacks.}
\label{tab:more_defenses}
\end{table}

Crucially, none of the defenses evaluated fully close the gap from the non-attacked model. This reflects a broader challenge in open-weight safety: defenses that suppress refusal behavior without eliminating underlying harmful knowledge will remain exploitable by simple jailbreaking attacks.

%% file: sections/relwork.tex

\textbf{Open-weight safety evaluation.} Because adversaries have direct access to open-weight model parameters, they can modify weights to remove safety behavior entirely~\citep{casper2025open}. These white-box attacks are an accessible and effective attack, as we discuss below. A growing line of work has developed defenses specifically designed to persist after adversarial modification. These defenses (i) train against simulated attacks~\citep{huang2024vaccine,liu2024targeted,huang2024booster,nguyen2026antibody,fan2025towards,tamirisa2025tamper,sanyal2025antidote}; (ii) make harmful fine-tuning induce capability collapse~\citep{henderson2023self,wang2025self,yi2025ctrap,chen2025sdd,feng2025token} or make its optimization landscape ill-conditioned~\citep{zheng2025model,rosati2025locking,rosati2026limits}; (iii) protect safety-relevant representations and parameters~\citep{rosati2024representation,zhao2025identifying,perin2025lox}; or (iv) modify the alignment data or procedure~\citep{liu2024buckle,liu2025pharmacist,chen2025vulnerability,cheng2025weaponization,yang2024preserving}.
TAR and SEAM (Section~\ref{sec:preliminaries}) are two representative safeguards that defend specifically against fine-tuning, and we additionally evaluate RepNoise, Vaccine, and Booster in Section~\ref{sec:defense}.

\textbf{Fine-tuning attacks.} Fine-tuning attacks pose a threat both to closed models via fine-tuning-as-a-service APIs and, more acutely, to open-weight models where adversaries have direct access to weights~\citep{huang2024harmful,casper2025open}. Fine-tuning attacks can break safety alignment by fine-tuning on a few harmful examples \citep{yang2023shadow,lermen2023lora,zhan2024removing,hu2024jogging}, fine-tuning on benign data \citep{qi2024fine,luckiadversarial,deeb2024unlearning,guo2026llms}, or simple perturbations to the pipeline~\citep{zhang2025exploring,qi2025evaluating}.

\textbf{Activation and weight-space attacks.} Safety behavior can also be bypassed via editing the model's activations or projecting model weights without fine-tuning. In the model's representation space, certain directions correspond to behaviors of interest; these directions can be added or subtracted at inference time to amplify or suppress these behaviors respectively~\citep{panickssery2023steering,lai2024projected,young2025comparative,zhao2025llms}. \citet{arditi2024refusal} leverage this insight and propose a simple search over layers, tokens, and scaling factors to maximize the effect of refusal suppression.

\textbf{Prompting and prefilling attacks.} Prompt-based jailbreaks require only black-box access and can elicit harmful outputs without modifying model weights \citep{zou2023universal}. Prefilling attacks are a more targeted variant which supply a non-refusal assistant prefix prior to generation~\citep{qi2025safety,andriushchenko2025jailbreaking,li2025prefill}. 

\textbf{Our work.} TAR and SEAM are designed to prevent adversarial misuse of open-weight models, but recent work has shown that they remain brittle to simple variations of fine-tuning attacks~\citep{zhang2025exploring,qi2025evaluating}, raising the question of whether they are robust to gradient-free attacks at all. While abliteration and prefilling are well-known attacks, prior work has not evaluated them against these safeguards, whose threat model assumes harmful capabilities must be introduced by fine-tuning. We show that these safeguards are vulnerable to abliteration and prefilling, and propose ART as a complementary defense that can be layered onto existing safeguards to reduce their vulnerability.

%% file: sections/conclusion.tex

Open-weight safeguards such as TAR and SEAM are designed to prevent adversarial misuse by resisting fine-tuning attacks. However, as shown in prior work and supported by our results, pretrained models contain significant amounts of harmful knowledge which cannot be easily removed via downstream fine-tuning safeguards. Thus, gradient-free attacks can simply elicit this knowledge without needing fine-tuning at all. To validate this, we show that Abliteration and Prefilling recover success rates of 16--97\% compared to below 10\% under the safeguard alone; 
the additional access available in the open-weight setting could be exploited to design stronger variants of both attacks.

As an initial step toward closing this gap, we propose ART, which defends against abliteration by simulating abliteration during training and performing gradient ascent on harmful outputs. While ART reduces attack success rates across all settings and can be layered onto TAR to improve robustness against both abliteration and fine-tuning attacks, no defense evaluated fully closes the gap to the no-attack baseline. This suggests that truly robust open-weight safety will require defenses that eliminate harmful knowledge rather than relying on refusal mechanisms. We hope our results motivate both stronger defenses and more comprehensive evaluation practices that consider both fine-tuning and gradient-free attacks in open-weight settings.

%% file: sections/appendix.tex
\textbf{Licenses and Attribution.} Our figures use images from Flaticon.com. As per their free license, we state that ``This cover has been designed using resources from Flaticon.com.''

\begin{figure}[h!]
    \centering
    \includegraphics[width=\linewidth]{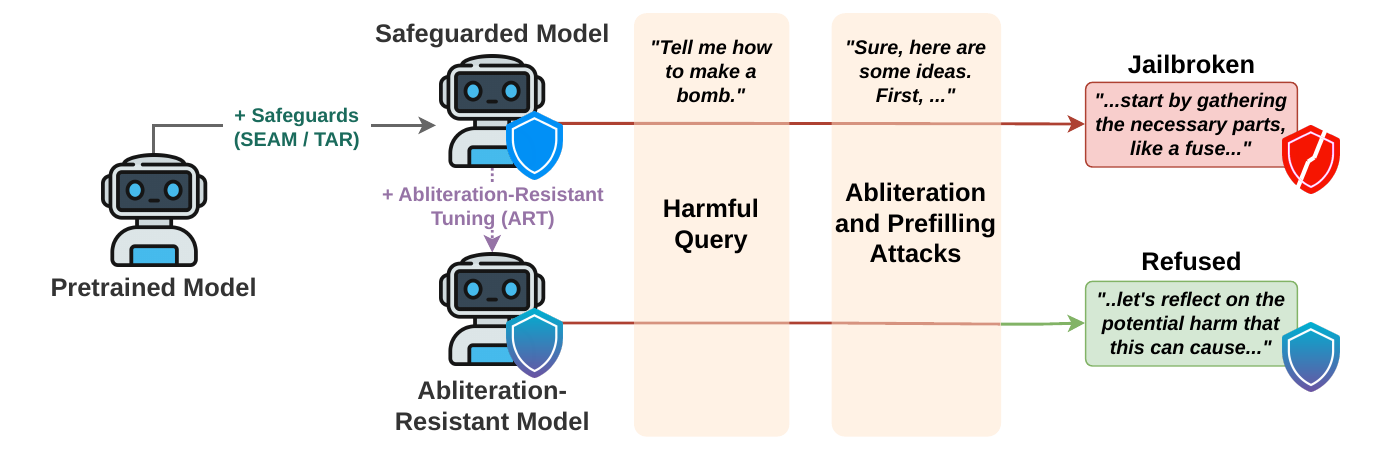}
    \caption{\textbf{Overview of our safeguard and attack pipeline.} A pretrained model is safeguarded with either TAR or SEAM; we then layer Abliteration-Resistant Tuning (ART) on top of these safeguards. We evaluate two inference-time attacks on the resulting models: \textit{abliteration}, which removes a refusal direction from the model's residual stream activations, and \textit{prefilling}, which bypasses refusal by injecting a compliant partial response (\textit{``Sure, here are some ideas. First, \ldots''}) before generation. Despite being trained to resist fine-tuning attacks, the safeguarded model remains vulnerable to inference-time attacks and produces harmful outputs (\textit{Jailbroken}), whereas the abliteration-resistant model continues to refuse (\textit{Refused}).}    \label{fig:placeholder}
\end{figure}

\subsection{Abliteration Attack Budget}\label{app:budget}

\textbf{The budget does not need to grow with model size.} In our main experiments, the full Abliteration attack searches over all layers, so larger models use more configurations. Table~\ref{tab:bestofn_sizes} shows that this is not necessary: with $N$=20 randomly sampled configurations, Abliteration recovers on average 64\% (43--76\%) of the full attack success rate for 1B models and 66\% (51--86\%) for 8B models. Thus, the number of configurations does not need to grow with the number of layers.

\begin{table*}[h]
\centering
\begin{tabular}{llcccccccccc}
\toprule
& & \multicolumn{5}{c}{\textbf{SEAM}} & \multicolumn{5}{c}{\textbf{TAR}} \\
\cmidrule(lr){3-7} \cmidrule(lr){8-12}
\textbf{Dataset} & \textbf{Size} & \textbf{$N$=1} & \textbf{5} & \textbf{10} & \textbf{20} & \textbf{Full} & \textbf{$N$=1} & \textbf{5} & \textbf{10} & \textbf{20} & \textbf{Full} \\
\midrule
\multirow{3}{*}{BeaverTails}
  & 1B & 3 & 12 & 19 & 28 & 43 & 1 & 4 & 6 & 7 & 10 \\
  & 3B & 2 &  8 & 13 & 19 & 41 & 2 & 7 & 9 & 13 & 22 \\
  & 8B & 4 & 16 & 26 & 39 & 73 & 5 & 12 & 15 & 19 & 26 \\
\midrule
\multirow{3}{*}{AdvBench}
  & 1B & 3 & 13 & 24 & 42 & 72 & 0 & 1 & 2 & 4 & 9 \\
  & 3B & 1 &  5 &  9 & 17 & 55 & 0 & 2 & 4 & 7 & 22 \\
  & 8B & 4 & 16 & 28 & 49 & 96 & 13 & 35 & 44 & 52 & 68 \\
\midrule
\multirow{3}{*}{HarmBench}
  & 1B & 3 & 14 & 24 & 38 & 59 & 1 & 5 & 8 & 12 & 16 \\
  & 3B & 1 &  6 & 10 & 18 & 50 & 2 & 9 & 14 & 19 & 34 \\
  & 8B & 5 & 22 & 36 & 56 & 93 & 30 & 52 & 59 & 66 & 77 \\
\bottomrule
\end{tabular}
\caption{Attack success rate (\%) of Abliteration with $N$ randomly sampled attack configurations across Llama model sizes. Full corresponds to the best-of-$N$ search over all layers and scaling factors ($N$=48, 84, and 96 for 1B, 3B, and 8B respectively). A fixed budget of $N$=20 recovers a similar fraction of the full attack success rate across sizes.}
\label{tab:bestofn_sizes}
\end{table*}

\subsection{Model Family Ablation}\label{app:families}

\textbf{Attacks are effective across model architecture.} Table~\ref{tab:asr_families} reports attack success rates on BeaverTails across three model families: Llama, Qwen, and Gemma. Across all families, Abliteration is a stronger attack than either baseline and prefill-only. Gemma is particularly vulnerable to Abliteration, reaching 93\% and 96\% under Abliteration and the combined attack respectively. One notable difference across families is the effect of Prefilling on Qwen SEAM, where Prefilling alone raises ASR from 1\% to 34\%, a much larger effect than seen in Llama or Gemma. Overall, the best attacks increase attack success rates over the Baseline by several times.

\begin{table*}[h!]
\centering
\begin{tabular}{llcccc}
\toprule
\textbf{Model Family} & \textbf{Safeguard} & \textbf{Baseline} & \textbf{Prefill} & \textbf{Abliterate} & \textbf{Abliterate + Prefill} \\
\midrule
\multirow{3}{*}{Llama}
  & Base  & 15 & 30 & 68 & \textbf{80} \\
  & SEAM  &  0 &  17 & 23 & \textbf{75} \\
  & TAR   &  4 &  3 & \textbf{16} & 11 \\
\midrule
\multirow{3}{*}{Qwen}
  & Base  & 30 & 27 & \textbf{70} & 68 \\
  & SEAM  &  1 &  34 & 16 & \textbf{63} \\
  & TAR   & 1 & 2 & 26 & \textbf{30} \\
\midrule
\multirow{3}{*}{Gemma}
  & Base  & 6 & 20 & 93 & \textbf{96} \\
  & SEAM  & 7 & 16 & \textbf{80} & 76 \\
  & TAR   & 5 & 4 & \textbf{76} & 66 \\
\bottomrule
\end{tabular}
\caption{Attack success rate (\%) on BeaverTails against 1-2B sized models across model families.}
\label{tab:asr_families}
\end{table*}

\subsection{Representation Analysis}\label{app:pca}

\textbf{TAR is more effective than SEAM.} While Abliteration and Prefilling significantly increase attack success rate against TAR (up to multiple times), TAR is substantially more resistant to attacks than Base or SEAM. In Table~\ref{tab:asr_8b}, the ASR on TAR ranges from as low as 9\% (BeaverTails, Prefill) up to 62\% (AdvBench, Abliterate). We hypothesize that this is because the objective of TAR minimizes harmful behavior more strongly through an inner meta-learning optimization, while the objective of SEAM combines vanilla gradient ascent on harmful data with a self-destructive loss. Below, we provide further intuition by visualizing the representations of each safeguard.

\begin{figure*}[h!]
\centering
\includegraphics[width=0.97\textwidth]{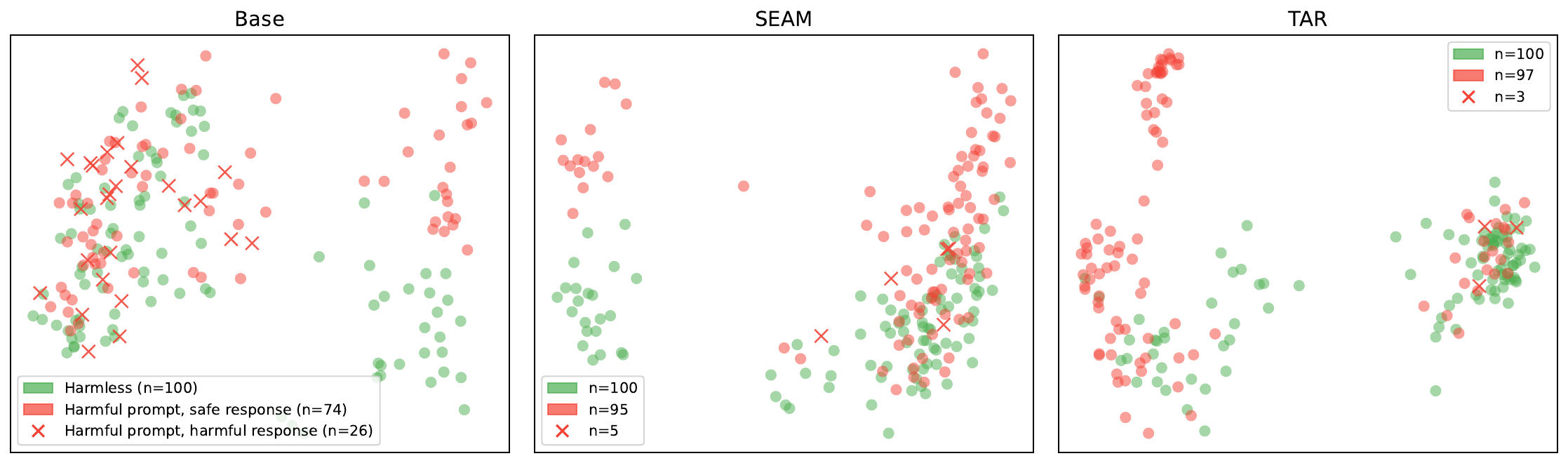}
\caption{PCA projections of residual stream activations (after generating 128 tokens) for Base, TAR, and SEAM models following Abliteration. Points represent harmful prompts that elicited a harmful response (×), harmful prompts that elicited a safe response (red), and harmless prompts (green). TAR shows clearer separation between harmful and harmless representations than SEAM, followed by Base, supporting the intuition that distinguishing between the two is essential for a strong safeguard.}
\label{fig:pca}
\end{figure*}

\textbf{TAR separates harmful and harmless representations more clearly.} In Figure~\ref{fig:pca}, we provide intuition on the effectiveness of SEAM and TAR compared to Base by visualizing the last-layer representations of each method. We compare last-token hidden states at layer 15 after abliterating the models and generating 128 tokens. TAR exhibits clearer separation between harmful and harmless inputs than SEAM, followed by Base. Because this is the same ranking we observe in terms of ascending ASR, this matches our expectation that more effective safeguards (i.e. TAR) exhibit a stronger ability to identify (and handle) harmful representations.

\subsection{ART Experimental Setup}\label{app:art_setup}
On Gemma-1B-IT, we fine-tune each defended model for an additional $200$ steps on the Anthropic HH dataset \citep{bai2022training}. After training, we repeat the same attack pipeline described in Section~\ref{sec:results}: we evaluate attack success rate on BeaverTails under the same four attack conditions and report benign utility on the Alpaca benchmark. ART introduces two additional hyperparameters, $\gamma_\text{ar}$ and $\gamma_\text{ret}$, which we set to $\gamma_\text{ar}=1$ and $\gamma_\text{ret}=2$ in all experiments. We train with a batch size of 4, 200 steps, and a learning rate of 2e-5. We recompute the optimal layer $l^*$ every 10 steps. While increasing the amount of data used and frequency for abliteration could potentially improve the robustness of ART, we intentionally did this to keep our setup lightweight.

When we benchmark fine-tuning attacks on Llama3.1-8B-Instruct (Table~\ref{tab:more_defenses}), we apply ART as described above with 1000 steps. We follow the harmful fine-tuning pipeline of \citet{huang2024vaccine} to evaluate all defenses, including ART: we fine-tune on 1000 samples containing 10\% harmful BeaverTails data and 90\% benign SST-2 data~\citep{socher2013recursive} for 20 epochs with a learning rate of 1e-5. We judge both fine-tuning and abliteration attacks with BeaverDam-7B~\citep{ji2023beavertails} on a held-out BeaverTails split. We additionally test multiple safeguards beyond SEAM and TAR: RepNoise~\citep{rosati2024representation}, Vaccine~\citep{huang2024vaccine}, and Booster~\citep{huang2024booster}.

%% file: rebuttal_refs.bib
@inproceedings{huang2024vaccine,
  title={{Vaccine}: Perturbation-aware Alignment for Large Language Models against Harmful Fine-tuning Attack},
  author={Huang, Tiansheng and Hu, Sihao and Liu, Ling},
  booktitle={Advances in Neural Information Processing Systems},
  year={2024}
}

@inproceedings{huang2024booster,
  title={{Booster}: Tackling Harmful Fine-tuning for Large Language Models via Attenuating Harmful Perturbation},
  author={Huang, Tiansheng and Hu, Sihao and Ilhan, Fatih and Tekin, Selim Furkan and Liu, Ling},
  booktitle={International Conference on Learning Representations},
  year={2025}
}

@article{liu2024targeted,
  title={{Targeted} Vaccine: Safety Alignment for Large Language Models against Harmful {Fine-Tuning} via Layer-wise Perturbation},
  author={Liu, Guozhi and Lin, Weiwei and Huang, Tiansheng and Mo, Ruichao and Mu, Qi and Shen, Li},
  journal={arXiv preprint arXiv:2410.09760},
  year={2024}
}

@inproceedings{hendrycks2021measuring,
  title={{Measuring} Massive Multitask Language Understanding},
  author={Hendrycks, Dan and Burns, Collin and Basart, Steven and Zou, Andy and Mazeika, Mantas and Song, Dawn and Steinhardt, Jacob},
  booktitle={International Conference on Learning Representations},
  year={2021}
}

@article{hughes2024best,
  title={{Best-of-N} Jailbreaking},
  author={Hughes, John and Price, Sara and Lynch, Aengus and Schaeffer, Rylan and Barez, Fazl and Koyejo, Sanmi and Sleight, Henry and Jones, Erik and Perez, Ethan and Sharma, Mrinank},
  journal={arXiv preprint arXiv:2412.03556},
  year={2024}
}

@article{chao2023jailbreaking,
  title={{Jailbreaking} Black Box Large Language Models in Twenty Queries},
  author={Chao, Patrick and Robey, Alexander and Dobriban, Edgar and Hassani, Hamed and Pappas, George J. and Wong, Eric},
  journal={arXiv preprint arXiv:2310.08419},
  year={2023}
}

@inproceedings{mehrotra2024tree,
  title={{Tree} of Attacks: Jailbreaking {Black-Box} {LLMs} Automatically},
  author={Mehrotra, Anay and Zampetakis, Manolis and Kassianik, Paul and Nelson, Blaine and Anderson, Hyrum and Singer, Yaron and Karbasi, Amin},
  booktitle={Advances in Neural Information Processing Systems},
  year={2024}
}

@inproceedings{nguyen2026antibody,
  title={{Antibody}: Strengthening Defense Against Harmful {Fine-Tuning} for Large Language Models via Attenuating Harmful Gradient Influence},
  author={Nguyen, Quoc Minh and Le, Trung and Wu, Jing and Bui, Anh Tuan and Harandi, Mehrtash},
  booktitle={International Conference on Learning Representations},
  year={2026}
}

@article{sanyal2025antidote,
  title={{AntiDote}: Bi-level Adversarial Training for {Tamper-Resistant} {LLMs}},
  author={Sanyal, Debdeep and Ray, Manodeep and Mandal, Murari},
  journal={arXiv preprint arXiv:2509.08000},
  year={2025}
}

@inproceedings{chen2025vulnerability,
  title={{Vulnerability-Aware} Alignment: Mitigating Uneven Forgetting in Harmful {Fine-Tuning}},
  author={Chen, Liang and Han, Xueting and Shen, Li and Bai, Jing and Wong, Kam-Fai},
  booktitle={International Conference on Machine Learning},
  year={2025}
}

@article{liu2025pharmacist,
  title={{Pharmacist}: Safety Alignment Data Curation for Large Language Models against Harmful Fine-tuning},
  author={Liu, Guozhi and Mu, Qi and Huang, Tiansheng and Wang, Xinhua and Shen, Li and Lin, Weiwei and Li, Zhang},
  journal={arXiv preprint arXiv:2510.10085},
  year={2025}
}

@article{perin2025lox,
  title={{LoX}: {Low-Rank} Extrapolation Robustifies {LLM} Safety Against Fine-tuning},
  author={Perin, Gabriel J. and Chen, Runjin and Chen, Xuxi and Hirata, Nina S. T. and Wang, Zhangyang and Hong, Junyuan},
  journal={arXiv preprint arXiv:2506.15606},
  year={2025}
}

@article{yi2025ctrap,
  title={{CTRAP}: Embedding Collapse Trap to Safeguard Large Language Models from Harmful {Fine-Tuning}},
  author={Yi, Biao and Huang, Tiansheng and Zhang, Baolei and Li, Tong and Nie, Lihai and Liu, Zheli and Shen, Li},
  journal={arXiv preprint arXiv:2505.16559},
  year={2025}
}

@inproceedings{zheng2025model,
  title={{Model} Immunization from a Condition Number Perspective},
  author={Zheng, Amber Yijia and Bai, Cedar Site and Bullins, Brian and Yeh, Raymond A.},
  booktitle={International Conference on Machine Learning},
  year={2025}
}

@inproceedings{chen2025sdd,
  title={{SDD}: {Self-Degraded} Defense against Malicious Fine-tuning},
  author={Chen, Zixuan and Lu, Weikai and Lin, Xin and Zeng, Ziqian},
  booktitle={Proceedings of the Annual Meeting of the Association for Computational Linguistics},
  year={2025}
}

@article{rosati2026limits,
  title={{Limits} of {Convergence-Rate} Control for {Open-Weight} Safety},
  author={Rosati, Domenic and Zeng, Xijie and Huang, Hong and Dionicio, Sebastian and Majumdar, Subhabrata and Rudzicz, Frank and Sajjad, Hassan},
  journal={arXiv preprint arXiv:2602.18868},
  year={2026}
}

@inproceedings{feng2025token,
  title={{Token} Buncher: Shielding {LLMs} from Harmful Reinforcement Learning {Fine-Tuning}},
  author={Feng, Weitao and Wang, Lixu and Lv, Peizhuo and Wei, Tianyi and Zhang, Jie and Gao, Chongyang and Zhan, Sinong and Dong, Wei},
  booktitle={Proceedings of the ACM Conference on Computer and Communications Security},
  year={2026}
}

@article{yu2023gptfuzzer,
  title={{GPTFUZZER}: Red Teaming Large Language Models with Auto-Generated Jailbreak Prompts},
  author={Yu, Jiahao and Lin, Xingwei and Yu, Zheng and Xing, Xinyu},
  journal={arXiv preprint arXiv:2309.10253},
  year={2023}
}

@inproceedings{socher2013recursive,
  title={Recursive Deep Models for Semantic Compositionality Over a Sentiment Treebank},
  author={Socher, Richard and Perelygin, Alex and Wu, Jean and Chuang, Jason and Manning, Christopher D. and Ng, Andrew and Potts, Christopher},
  booktitle={Proceedings of the Conference on Empirical Methods in Natural Language Processing},
  year={2013}
}

@inproceedings{liu2024buckle,
  title={Buckle Up: Robustifying {LLMs} at Every Customization Stage via Data Curation},
  author={Liu, Xiaoqun and Liang, Jiacheng and Tang, Luoxi and You, Chenyu and Ye, Muchao and Xi, Zhaohan},
  booktitle={International Conference on Learning Representations},
  year={2025}
}

@inproceedings{zhao2025identifying,
  title={Identifying and Tuning Safety Neurons in Large Language Models},
  author={Zhao, Yiran and Zhang, Wenxuan and Xie, Yuxi and Goyal, Anirudh and Kawaguchi, Kenji and Shieh, Michael Qizhe},
  booktitle={International Conference on Learning Representations},
  year={2025}
}

@inproceedings{yang2024preserving,
  title={Preserving Safety in Fine-Tuned Large Language Models: A Systematic Evaluation and Mitigation Strategy},
  author={Yang, Tsung-Huan and Huang, Ko-Wei and Li, Yung-Hui and Ku, Lun-Wei},
  booktitle={NeurIPS Workshop on Safe Generative AI},
  year={2024}
}

@inproceedings{rosati2025locking,
  title={Locking Open Weight Models with Spectral Deformation},
  author={Rosati, Domenic and Dionicio, Sebastian and Zeng, Xijie and Majumdar, Subhabrata and Rudzicz, Frank and Sajjad, Hassan},
  booktitle={International Conference on Machine Learning},
  year={2025}
}

@inproceedings{cheng2025weaponization,
  title={On Weaponization-Resistant Large Language Models with Prospect Theoretic Alignment},
  author={Cheng, Zehua and Zhang, Manying and Sun, Jiahao and Dai, Wei},
  booktitle={Proceedings of the International Conference on Computational Linguistics},
  pages={10309--10324},
  year={2025}
}


%% file: references.bib
@article{wallace2025estimating,
  title={Estimating worst-case frontier risks of open-weight llms},
  author={Wallace, Eric and Watkins, Olivia and Wang, Miles and Chen, Kai and Koch, Chris},
  journal={arXiv preprint arXiv:2508.03153},
  year={2025}
}

@article{minaee2024large,
  title={Large language models: A survey},
  author={Minaee, Shervin and Mikolov, Tomas and Nikzad, Narjes and Chenaghlu, Meysam and Socher, Richard and Amatriain, Xavier and Gao, Jianfeng},
  journal={arXiv preprint arXiv:2402.06196},
  year={2024}
}

@article{taori2023alpaca,
  title={Alpaca: A strong, replicable instruction-following model},
  author={Taori, Rohan and Gulrajani, Ishaan and Zhang, Tianyi and Dubois, Yann and Li, Xuechen and Guestrin, Carlos and Liang, Percy and Hashimoto, Tatsunori B},
  journal={Stanford Center for Research on Foundation Models. https://crfm. stanford. edu/2023/03/13/alpaca. html},
  volume={3},
  number={6},
  pages={7},
  year={2023}
}

@article{ouyang2022training,
  title={Training language models to follow instructions with human feedback},
  author={Ouyang, Long and Wu, Jeffrey and Jiang, Xu and Almeida, Diogo and Wainwright, Carroll and Mishkin, Pamela and Zhang, Chong and Agarwal, Sandhini and Slama, Katarina and Ray, Alex and others},
  journal={Advances in neural information processing systems},
  volume={35},
  pages={27730--27744},
  year={2022}
}

@article{rafailov2023direct,
  title={Direct preference optimization: Your language model is secretly a reward model},
  author={Rafailov, Rafael and Sharma, Archit and Mitchell, Eric and Manning, Christopher D and Ermon, Stefano and Finn, Chelsea},
  journal={Advances in neural information processing systems},
  volume={36},
  pages={53728--53741},
  year={2023}
}

@inproceedings{zhao2025llms,
  title={LLMs Encode Harmfulness and Refusal Separately},
  author={Zhao, Jiachen and Huang, Jing and Wu, Zhengxuan and Bau, David and Shi, Weiyan},
  booktitle={The Thirty-ninth Annual Conference on Neural Information Processing Systems},
  year={2025}
}

@inproceedings{tamirisa2025tamper,
  title={Tamper-Resistant Safeguards for Open-Weight {LLM}s},
  author={Tamirisa, Rishub and Bharathi, Bhrugu and Phan, Long and Zhou, Andy and Gatti, Alice and Suresh, Tarun and Lin, Maxwell and Wang, Justin and Wang, Rowan and Arel, Ron and others},
  booktitle={The Thirteenth International Conference on Learning Representations},
  year={2025}
}

@inproceedings{henderson2023self,
  title={Self-Destructing Models: Increasing the Costs of Harmful Dual Uses of Foundation Models},
  author={Henderson, Peter and Mitchell, Eric and Manning, Christopher and Jurafsky, Dan and Finn, Chelsea},
  booktitle={Proceedings of the 2023 AAAI/ACM Conference on AI, Ethics, and Society},
  pages={287--296},
  year={2023}
}

@article{wang2025self,
  title={Self-Destructive Language Model},
  author={Wang, Yuhui and Zhu, Rongyi and Wang, Ting},
  journal={arXiv preprint arXiv:2505.12186},
  year={2025}
}

@article{rosati2024representation,
  title={Representation Noising: A Defence Mechanism Against Harmful Finetuning},
  author={Rosati, Domenic and Wehner, Jan and Williams, Kai and Bartoszcze, {\L}ukasz and Atanasov, David and Gonzales, Robie and Majumdar, Subhabrata and Maple, Carsten and Sajjad, Hassan and Rudzicz, Frank},
  journal={Advances in Neural Information Processing Systems},
  volume={37},
  pages={12636--12676},
  year={2024}
}

@inproceedings{fan2025towards,
  title={Towards LLM Unlearning Resilient to Relearning Attacks: A Sharpness-Aware Minimization Perspective and Beyond},
  author={Fan, Chongyu and Jia, Jinghan and Zhang, Yihua and Ramakrishna, Anil and Hong, Mingyi and Liu, Sijia},
  booktitle={International Conference on Machine Learning},
  pages={15762--15778},
  year={2025},
  organization={PMLR}
}

@article{bai2022training,
  title={Training a helpful and harmless assistant with reinforcement learning from human feedback},
  author={Bai, Yuntao and Jones, Andy and Ndousse, Kamal and Askell, Amanda and Chen, Anna and DasSarma, Nova and Drain, Dawn and Fort, Stanislav and Ganguli, Deep and Henighan, Tom and others},
  journal={arXiv preprint arXiv:2204.05862},
  year={2022}
}

@phdthesis{zhang2025exploring,
  title={Exploring Fine-Tuning Techniques for Removing Tamper-Resistant Safeguards for Open-Weight LLMs},
  author={Zhang, Sarah},
  year={2025},
  school={Massachusetts Institute of Technology}
}

@inproceedings{qi2025evaluating,
  title={On Evaluating the Durability of Safeguards for Open-Weight {LLM}s},
  author={Qi, Xiangyu and Wei, Boyi and Carlini, Nicholas and Huang, Yangsibo and Xie, Tinghao and He, Luxi and Jagielski, Matthew and Nasr, Milad and Mittal, Prateek and Henderson, Peter},
  booktitle={The Thirteenth International Conference on Learning Representations},
  year={2025}
}

@article{luckiadversarial,
  title={An Adversarial Perspective on Machine Unlearning for {AI} Safety},
  author={{\L}ucki, Jakub and Wei, Boyi and Huang, Yangsibo and Henderson, Peter and Tram{\`e}r, Florian and Rando, Javier},
  journal={Transactions on Machine Learning Research},
  year={2024}
}

@article{deeb2024unlearning,
  title={Do Unlearning Methods Remove Information from Language Model Weights?},
  author={Deeb, Aghyad and Roger, Fabien},
  journal={arXiv preprint arXiv:2410.08827},
  year={2024}
}

@article{casper2025open,
  title={Open technical problems in open-weight AI model risk management},
  author={Casper, Stephen and O'Brien, Kyle and Longpre, Shayne and Seger, Elizabeth and Klyman, Kevin and Bommasani, Rishi and Nrusimha, Aniruddha and Shumailov, Ilia and Mindermann, S{\"o}ren and Basart, Steven and others},
  journal={Transactions on Machine Learning Research},
  year={2025}
}

@article{ji2023beavertails,
  title={Beavertails: Towards improved safety alignment of llm via a human-preference dataset},
  author={Ji, Jiaming and Liu, Mickel and Dai, Josef and Pan, Xuehai and Zhang, Chi and Bian, Ce and Chen, Boyuan and Sun, Ruiyang and Wang, Yizhou and Yang, Yaodong},
  journal={Advances in Neural Information Processing Systems},
  volume={36},
  pages={24678--24704},
  year={2023}
}

@article{zou2023universal,
  title={Universal and transferable adversarial attacks on aligned language models},
  author={Zou, Andy and Wang, Zifan and Carlini, Nicholas and Nasr, Milad and Kolter, J Zico and Fredrikson, Matt},
  journal={arXiv preprint arXiv:2307.15043},
  year={2023}
}

@article{mazeika2024harmbench,
  title={Harmbench: A standardized evaluation framework for automated red teaming and robust refusal},
  author={Mazeika, Mantas and Phan, Long and Yin, Xuwang and Zou, Andy and Wang, Zifan and Mu, Norman and Sakhaee, Elham and Li, Nathaniel and Basart, Steven and Li, Bo and others},
  journal={arXiv preprint arXiv:2402.04249},
  year={2024}
}

@article{arditi2024refusal,
  title={Refusal in Language Models Is Mediated by a Single Direction},
  author={Arditi, Andy and Obeso, Oscar and Syed, Aaquib and Paleka, Daniel and Panickssery, Nina and Gurnee, Wes and Nanda, Neel},
  journal={Advances in Neural Information Processing Systems},
  volume={37},
  pages={136037--136083},
  year={2024}
}

@misc{lai2024projected,
  title={Projected Abliteration},
  author={Lai, Grim Jim},
  year={2024},
  howpublished={\url{https://huggingface.co/blog/grimjim/projected-abliteration}},
  note={Hugging Face Blog}
}

@article{young2025comparative,
  title={Comparative Analysis of {LLM} Abliteration Methods: A Cross-Architecture Evaluation},
  author={Young, Richard J},
  journal={arXiv preprint arXiv:2512.13655},
  year={2025}
}

@article{panickssery2023steering,
  title={Steering llama 2 via contrastive activation addition},
  author={Panickssery, Nina and Gabrieli, Nick and Schulz, Julian and Tong, Meg and Hubinger, Evan and Turner, Alexander Matt},
  journal={arXiv preprint arXiv:2312.06681},
  year={2023}
}

@inproceedings{andriushchenko2025jailbreaking,
  title={Jailbreaking Leading Safety-Aligned LLMs with Simple Adaptive Attacks},
  author={Andriushchenko, Maksym and Croce, Francesco and Flammarion, Nicolas},
  booktitle={The Thirteenth International Conference on Learning Representations},
  year={2025}
}

@inproceedings{vega2024bypassing,
  title={Bypassing the Safety Training of Open-Source LLMs with Priming Attacks},
  author={Vega, Jason and Chaudhary, Isha and Xu, Changming and Singh, Gagandeep},
  booktitle={The Second Tiny Papers Track at ICLR 2024},
  year={2024}
}

@inproceedings{qi2025safety,
  title={Safety Alignment Should Be Made More Than Just a Few Tokens Deep},
  author={Qi, Xiangyu and Panda, Ashwinee and Lyu, Kaifeng and Ma, Xiao and Roy, Subhrajit and Beirami, Ahmad and Mittal, Prateek and Henderson, Peter},
  booktitle={The Thirteenth International Conference on Learning Representations},
  year={2025}
}

@article{li2025prefill,
  title={Prefill-Level Jailbreak: A Black-Box Risk Analysis of Large Language Models},
  author={Li, Yakai and Hu, Jiekang and Sang, Weiduan and Ma, Luping and Nie, Dongsheng and Zhang, Weijuan and Yu, Aimin and Su, Yi and Huang, Qingjia and Zhou, Qihang},
  journal={arXiv preprint arXiv:2504.21038},
  year={2025}
}

@article{guo2026llms,
  title={LLMs Can Unlearn Refusal with Only 1,000 Benign Samples},
  author={Guo, Yangyang and Xu, Ziwei and Liu, Si and Zheng, Zhiming and Kankanhalli, Mohan},
  journal={arXiv preprint arXiv:2601.19231},
  year={2026}
}

@inproceedings{qi2024fine,
  title={Fine-Tuning Aligned Language Models Compromises Safety, Even When Users Do Not Intend To!},
  author={Qi, Xiangyu and Zeng, Yi and Xie, Tinghao and Chen, Pin-Yu and Jia, Ruoxi and Mittal, Prateek and Henderson, Peter},
  booktitle={The Twelfth International Conference on Learning Representations},
  year={2024}
}

@article{yang2023shadow,
  title={Shadow alignment: The ease of subverting safely-aligned language models},
  author={Yang, Xianjun and Wang, Xiao and Zhang, Qi and Petzold, Linda and Wang, William Yang and Zhao, Xun and Lin, Dahua},
  journal={arXiv preprint arXiv:2310.02949},
  year={2023}
}

@inproceedings{yi2024vulnerability,
  title={On the vulnerability of safety alignment in open-access llms},
  author={Yi, Jingwei and Ye, Rui and Chen, Qisi and Zhu, Bin and Chen, Siheng and Lian, Defu and Sun, Guangzhong and Xie, Xing and Wu, Fangzhao},
  booktitle={Findings of the Association for Computational Linguistics: ACL 2024},
  pages={9236--9260},
  year={2024}
}

@article{lermen2023lora,
  title={Lora fine-tuning efficiently undoes safety training in llama 2-chat 70b},
  author={Lermen, Simon and Rogers-Smith, Charlie and Ladish, Jeffrey},
  journal={arXiv preprint arXiv:2310.20624},
  year={2023}
}

@inproceedings{zhan2024removing,
  title={Removing rlhf protections in gpt-4 via fine-tuning},
  author={Zhan, Qiusi and Fang, Richard and Bindu, Rohan and Gupta, Akul and Hashimoto, Tatsunori B and Kang, Daniel},
  booktitle={Proceedings of the 2024 Conference of the North American Chapter of the Association for Computational Linguistics: Human Language Technologies (Volume 2: Short Papers)},
  pages={681--687},
  year={2024}
}

@inproceedings{hu2024jogging,
  title={Jogging the memory of unlearned models through targeted relearning attacks},
  author={Hu, Shengyuan and Fu, Yiwei and Wu, Steven and Smith, Virginia},
  booktitle={ICML 2024 Workshop on Foundation Models in the Wild},
  year={2024}
}

@article{huang2024harmful,
  title={Harmful fine-tuning attacks and defenses for large language models: A survey},
  author={Huang, Tiansheng and Hu, Sihao and Ilhan, Fatih and Tekin, Selim Furkan and Liu, Ling},
  journal={arXiv preprint arXiv:2409.18169},
  year={2024}
}
